\documentclass[amstex,12pt,thmsa,sw20bams]{article}%
\usepackage{amsfonts}
\usepackage[utf8]{inputenc}
\usepackage{graphicx}
\usepackage[francais]{babel}
\usepackage{amsmath}
\usepackage{amssymb}%
\providecommand{\U}[1]{\protect\rule{.1in}{.1in}}
\ifx\pdfoutput\relax\let\pdfoutput=\undefined\fi
\newcount\msipdfoutput
\ifx\pdfoutput\undefined\else
\ifcase\pdfoutput\else
\ifx\paperwidth\undefined\else
\ifdim\paperheight=0pt\relax\else\pdfpageheight\paperheight\fi
\ifdim\paperwidth=0pt\relax\else\pdfpagewidth\paperwidth\fi
\fi\fi\fi
\begin{document}

\author{Choulakian V and Allard J, Universit\'{e} de Moncton, Canada
\and vartan.choulakian@umoncton.ca, jacques.allard@umoncton.ca}
\title{Visualization of Extremely Sparse Contingency Table by Taxicab Correspondence
Analysis: A Case Study of Textual Data }
\date{August 2023}
\maketitle

\begin{abstract}
We present an overview of taxicab correspondence analysis, a robust variant of
correspondence analysis, for visualization of extremely sparse contingency
tables. In particular we visualize an extremely sparse textual data set of
size 590 by 8265 concerning fragments of 8 sacred books recently introduced by
Sah and Fokou\'{e} (2019) and studied quite in detail by (12 + 1) dimension
reduction methods (t-SNE, UMAP, PHATE,...) by Ma, Sun and Zou (2022).



\end{abstract}

\section{\textbf{Introduction}}

Extremely sparse contingency tables often are high-dimensional and ubiquitous
in many scientific disciplines; their embeddings into low-dimensional spaces
for visualization is an active area of research. This article offers a study
from the perspective of correspondence analysis (CA) and its robust variant
taxicab CA (TCA). We start by summarizing the results of the main paper which
motivated our work.

An extremely sparse contingency table, textual data set or a bag of words
recently introduced by Sah and Fokou\'{e} (2019), of size $I\times
J=590\times8265$, where $I=590$ represents fragments of chapters of 8 sacred
books and $J=8265$ represents the number of distinct words, is
studied-visualized quite in detail by (12+1) dimension reduction methods in
Ma, Sun and Zou (2022,2023). The (12+1) methods are:\bigskip

1) Principal component analysis (PCA) (Hotelling 1933)

2) Multi-dimensional scaling (MDS) (Torgerson 1952 and Gower 1966)

3) Non-metric MDS (iMDS) (Kruskal 1964)

4) Sammon's mapping (Sammon) (Sammon 1969)

5) Kernel PCA (kPCA) (Sch\"{o}lkopf, Smola and M\"{u}ller 1998)

6) Locally linear embedding (LLE) (Roweis and Saul 2000)

7) Isomap (Tenenbaum, Silva and Langford 2000)

8) Hessian LLE (HLLE) (Donoho and Grimes 2003)

9) Laplacian eigenmap (LEIM) (Belkin and Niyogi 2003)

10) t-SNE (van der Maaten and Hinton 2008)

11) Uniform manifold approximation and projection for dimension reduction
(UMAP) (McInnes, Healy and Melville 2018)

12) (PHATE) (Moon et al. 2019)

13) Meta-visualization (meta-spec) (Ma, Sun and Zou 2022, 2023)

\subsection{\textbf{Remarks on the 13 maps}}

1) Figure 10 in (Ma, Sun and Zou 2022) shows that PCA and MDS maps are
identical, for MDS is the dual of PCA, a well known result, see Torgerson
(1952) and Gower (1966).

2) For the four methods, kPCA, t-SNE, UMAP and PHATE, two tuning parameters
are used, so there are two maps designated by the addition of integers either
1 or 2 on the title of the maps. This means that the last method
meta-visualization named (meta-spec) is based on the 16 candidate Euclidean
distances computed from the obtained 2-dimensional maps.

3) Only\ 12 maps are reproduced in (Ma, Sun and Zou 2022)'s paper: PHATE2,
t-SNE2, kPCA1 and meta-spec in Figure 3; Sammon, LLE, LEIM, UMAP, PCA, MDS,
ISOMAP, PHATE in Figure 10.

4) Maps produced by iMDS and HLLE are missing. Furthermore, two almost
identical maps produced by PHATE are represented in Figures 3 and 10.

5) The map (meta-spec) is the best: it discriminates completely the Eastern
and the Western (biblical) chapters; while only the 2 PHATE maps somewhat
ambiguously discriminate the two groups. We will see that the TCA map
discriminates the two groups almost as well as the (meta-spec) map;
furthermore it provides additional complementary information on the two groups
by the interpretation of words associated with each group, as is practised in CA.

6) All of the above mentioned methods are based on the pairwise Euclidean
distances among the 590 chapters, where each chapter is a vector of counts of
8260 words. The well-known phenomenon of the norm concentration of the
Euclidean distances, see for instance among others (Fran\c{c}ois, Wertz,
Verleysen 2007) and (Lee and Verheysen 2011), manifests itself clearly and
differently in the high-dimensional data space and in the embedded
low-dimensional visualization maps: this is the main reason that the maps
produced by the 12 simple methods are not sufficiently good enough.

\subsection{Contents of this paper}

We see from the above summary that neither correspondence analysis (CA) nor
its robust $l_{1}$ variant named taxicab CA (TCA) have been applied for
visualization. So, our primary goal is to show that for sparse, and in
particular for extremely sparse high-dimensional contingency tables it is
worth applying CA and TCA jointly. For small sized sparse data sets see
Mallet-Gauthier and Choulakian (2015) and Choulakian (2017).

Benz\'{e}cri (1973) is the reference book on CA\ but difficult to read because
of his use of tensor notation. Among others, Greenacre (1984) is the best
representative of Benz\'{e}cri's approach; Beh and Lombardo (2014) present a
panoramic review of CA.

We suppose that the theory of CA is known. So the contents of this paper are:
In section 2 we present an overview of taxicab singular value decomposition
(TSVD) on which TCA is developed; in section 3, we present an overview of TCA.
In section 4, we analyze the religious data set introduced above; finally we
conclude in section 5.

\section{An\ overview\ of\ taxicab singular value decomposition}

Consider a matrix $\mathbf{X}$\ of size $I\times J$ and $rank(\mathbf{X}%
)=k$\textbf{.} Taxicab singular value decomposition (TSVD) of \textbf{X} is a
decomposition similar to SVD of \textbf{X}, see Choulakian (2006, 2016).

For a vector $\mathbf{u}=\mathbf{(}u_{i}),$ its taxicab or L$_{1}$ norm is
$\left\vert \left\vert \mathbf{u}\right\vert \right\vert _{1}=\sum_{i}%
|u_{i}|,$ the Euclidean or L$_{2}$ norm is $\left\vert \left\vert
\mathbf{u}\right\vert \right\vert _{2}=(\sum_{i}|u_{i}|^{2})^{1/2}$ and the
L$_{\infty}$ norm is $\left\vert \left\vert \mathbf{u}\right\vert \right\vert
_{\infty}=\max_{i}\ |u_{i}|$.

In TSVD the calculation of the dispersion measures $(\delta_{\alpha})$,
principal axes ($\mathbf{u}_{\alpha},\mathbf{v}_{\alpha})$ and principal
scores $(\mathbf{a}_{\alpha},\mathbf{b}_{\alpha})$ for $\alpha=1,...,k$ is
done in a stepwise manner. We put $\mathbf{X}_{1}=\mathbf{X}=(x_{ij})$ and
$\mathbf{X_{\alpha}}$ be the residual matrix at the $\alpha$-th iteration for
$\alpha=1,...,k$.

The variational definitions of the TSVD at the $\alpha$-th iteration are%

\begin{align}
\delta_{\alpha}  &  =\max_{\mathbf{u\in%
\mathbb{R}
}^{J}}\frac{\left\vert \left\vert \mathbf{X_{\alpha}u}\right\vert \right\vert
_{1}}{\left\vert \left\vert \mathbf{u}\right\vert \right\vert _{\infty}}%
=\max_{\mathbf{v\in%
\mathbb{R}
}^{I}}\ \frac{\left\vert \left\vert \mathbf{X_{\alpha}^{\prime}v}\right\vert
\right\vert _{1}}{\left\vert \left\vert \mathbf{v}\right\vert \right\vert
_{\infty}}=\max_{\mathbf{u\in%
\mathbb{R}
}^{J},\mathbf{v\in%
\mathbb{R}
}^{I}}\frac{\mathbf{v}^{\prime}\mathbf{X_{\alpha}u}}{\left\vert \left\vert
\mathbf{u}\right\vert \right\vert _{\infty}\left\vert \left\vert
\mathbf{v}\right\vert \right\vert _{\infty}},\tag{1}\\
&  =\max\ ||\mathbf{X_{\alpha}u||}_{1}\ \ \text{subject to }\mathbf{u}%
\in\left\{  -1,+1\right\}  ^{J},\nonumber\\
&  =\max\ ||\mathbf{X_{\alpha}^{\prime}v||}_{1}\ \ \text{subject to
}\mathbf{v}\in\left\{  -1,+1\right\}  ^{I},\nonumber\\
&  =\max\mathbf{v}^{\prime}\mathbf{X_{\alpha}u}\text{ \ subject to
\ }\mathbf{u}\in\left\{  -1,+1\right\}  ^{J},\mathbf{v}\in\left\{
-1,+1\right\}  ^{I}.\nonumber
\end{align}
The $\alpha$-th principal axes are%
\begin{equation}
\mathbf{u}_{\alpha}\ =\arg\max_{\mathbf{u}\in\left\{  -1,+1\right\}  ^{J}%
}\left\vert \left\vert \mathbf{X_{\alpha}u}\right\vert \right\vert _{1}\text{
\ \ and \ \ }\mathbf{v}_{\alpha}\ =\arg\max_{\mathbf{v}\in\left\{
-1,+1\right\}  ^{I}}\left\vert \left\vert \mathbf{X_{\alpha}v}\right\vert
\right\vert _{1}\text{,} \tag{2}%
\end{equation}
and the $\alpha$-th principal projections of the rows and the columns are
\begin{equation}
\mathbf{a}_{\alpha}=\mathbf{X_{\alpha}u}_{\alpha}\text{ \ and \ }%
\mathbf{b}_{\alpha}=\mathbf{X_{\alpha}^{\prime}v}_{\alpha}. \tag{3}%
\end{equation}
Furthermore, the following relations are also useful%
\begin{equation}
\mathbf{u}_{\alpha}=sign(\mathbf{b}_{\alpha})\text{ \ and \ }\mathbf{v}%
_{\alpha}=sign(\mathbf{a}_{\alpha}), \tag{4}%
\end{equation}
where $sign(.)$ is the coordinatewise sign function, $sign(x)=1$ \ if \ $x>0,
$ \ and \ $sign(x)=-1$ \ if \ $x\leq0.$

The $\alpha$-th taxicab dispersion measure $\delta_{\alpha}$ can be
represented in many different ways%
\begin{align}
\delta_{\alpha}\  &  =\left\vert \left\vert \mathbf{X_{\alpha}u}_{\alpha
}\right\vert \right\vert _{1}=\left\vert \left\vert \mathbf{a}_{\alpha
}\right\vert \right\vert _{1}=\mathbf{a}_{\alpha}^{\prime}\mathbf{v}_{\alpha
},\tag{5}\\
&  =\left\vert \left\vert \mathbf{X_{\alpha}^{\prime}v}_{\alpha}\right\vert
\right\vert _{1}=\left\vert \left\vert \mathbf{b}_{\alpha}\right\vert
\right\vert _{1}=\mathbf{b}_{\alpha}^{\prime}\mathbf{u}_{\alpha}\nonumber\\
&  =\mathbf{v}_{\alpha}{}^{\prime}\mathbf{X_{\alpha}u}_{\alpha}%
=sign(\mathbf{a}_{\alpha})^{\prime}\mathbf{X_{\alpha}}sign(\mathbf{b}_{\alpha
}).\nonumber
\end{align}
The $(\alpha+1)$-th residual matrix is
\begin{equation}
\mathbf{X_{\alpha+1}}=\mathbf{X_{\alpha}-a}_{\alpha}\mathbf{b}_{\alpha
}^{\prime}/\delta_{\alpha}. \tag{6}%
\end{equation}
An interpretation of the term $\mathbf{a}_{\alpha}\mathbf{b}_{\alpha}^{\prime
}/\delta_{\alpha}$ in (6) is that, it represents the best rank-1 approximation
of the residual matrix $\mathbf{X_{\alpha}}$, in the sense of the taxicab
matrix norm (1).

Thus TSVD of $\mathbf{X}$ corresponds to the bilinear decomposition%

\begin{equation}
x_{ij}=\sum_{\alpha=1}^{k}a_{\alpha}(i)b_{\alpha}(j)/\delta_{\alpha}, \tag{7}%
\end{equation}
a decomposition similar to SVD, but where the vectors $(\mathbf{a}_{\alpha
},\mathbf{b}_{\alpha})$ for $\alpha=1,...,k$ are conjugate; that is%
\begin{align}
\mathbf{a}_{\alpha}^{\prime}\mathbf{v}_{\beta}  &  =\mathbf{a}_{\alpha
}^{\prime}sign(\mathbf{a}_{\beta})\tag{8}\\
&  =\mathbf{b}_{\alpha}^{\prime}\mathbf{u}_{\beta}=\mathbf{b}_{\alpha}%
^{\prime}sign(\mathbf{b}_{\beta})\nonumber\\
&  =0\text{ for }\alpha\geq\beta+1.\nonumber
\end{align}

In the package TaxicabCA in R, the calculation of the principal component
weights, $\mathbf{u}_{\alpha}$ and $\mathbf{v}_{\alpha},$ are accomplished by
three algorithms. The first one, based on complete enumeration equation (2),
is named \textit{exhaustive}. The second one, based on iterating the
transition formulae (3,4), is named \textit{criss-cross}. The third one, based
on the genetic algorithm, is named \textit{genetic}.

\section{Taxicab correspondence analysis: An overview}

CA is a dimension reduction method for exploratory visualization of
contingency tables based on Benz\'{e}cri's principle of distributional
equivalence property.

In the journal \textit{Revue Philosophique de la France et de l'\'{E}tranger,
}Benz\'{e}cri (1966) wrote an elaborate article entitled \textit{Linguistique
et math\'{e}matique}, where he presented his project of uncovering grammatical
rules or patterns from texts by announcing his conceptual formulation of the
principle of distributional equivalence \textquotedblright\textit{que des
\'{e}l\'{e}ments distributionnellement proches soient proches sur le diagramme
et r\'{e}ciproquement}\textquotedblright\ on which the geometric-mathematical
formulation of CA was essentially developed, and which can be described by the
following two definitions.\bigskip

\textbf{Definition 1: }Two non negative vectors \textbf{x} and \textbf{y} of
the same size are distributionally equivalent if $\mathbf{x=}C\mathbf{y}$,
where $C>0.\bigskip$

\textbf{Definition 2}: A method of analysis satisfies distributional
equivalence property, if it is applied to a table of nonnegative values where
two rows ( or two columns) are proportional, then the method will produce maps
where the two proportional rows (or the two proportional columns) coincide.
Furthermore, one can merge the proportional rows (or the proportional
columns), and the method will keep the results invariant.\bigskip

\textbf{Remark 1: }Many different methods satisfy Definition 2, see among
others Choulakian, Allard and Mahdi (2023).

\subsection{Preliminaries}

Let $\mathbf{N}=(n_{ij})$\textbf{ }for $i=1,...,I$ and $j=1,...,J$ be a 2-way
contingency table and $\mathbf{P=N/}n=(p_{ij})$ of size $I\times J$ the
associated correspondence matrix (probability table) of \textbf{N}, where
$n=\sum_{i,j}n_{ij}$. We define as usual $p_{i+}=\sum_{j=1}^{J}p_{ij}$ the
marginal probability of the $i$-th row, and similarly $p_{+j}=\sum_{i=1}%
^{I}p_{ij}$ the marginal property of the $j$-th column.

The CA association index of \textbf{P} is%

\begin{align*}
\Delta_{ij}  &  =(\frac{p_{ij}}{p_{i+}p_{+j}}-1)\\
&  =\frac{1}{p_{i+}p_{+j}}(p_{ij}-p_{i+}p_{+j})\\
&  =\frac{1}{p_{i+}}(\frac{p_{ij}}{p_{+j}}-p_{i+})\\
&  =\frac{1}{p_{+j}}(\frac{p_{ij}}{p_{i+}}-p_{+j})
\end{align*}

where: $\frac{p_{ij}}{p_{i+}p_{+j}}$ is density function of $p_{ij}$ with
respect to $p_{i+}p_{+j},$ $\frac{p_{ij}}{p_{+j}}$ is the profile of the
$j$-th column and similarly $\frac{p_{ij}}{p_{i+}}$ is the profile of the
$i$-th row.

The CA or Taxicab CA (TCA) decomposition is%

\begin{equation}
\frac{p_{ij}}{p_{i+}p_{+j}}-1=\sum_{\alpha=1}^{k}f_{\alpha}(i)g_{\alpha
}(j)/\delta_{\alpha}, \tag{9}%
\end{equation}
where for principal dimensions $\alpha=1,...,k$, $\delta_{\alpha}$ is the
dispersion measure and $(f_{\alpha}(i),g_{\alpha}(j))$ are the principal
factor scores of the rows and the columns respectively; and they are
calculated by generalized singular value decomposition (SVD) or taxicab SVD
(TSVD) of $\Delta_{ij}$, where $k=rank(\Delta_{ij}).$\bigskip

In CA, the parameters $(f_{\alpha}(i),g_{\alpha}(j),\delta_{\alpha})$ in (9)
satisfy: for $\alpha,\beta=1,...,k$

$\delta_{\alpha}^{2}=\sum_{\alpha=1}^{k}f_{\alpha}^{\ \ 2}(i)p_{i+}%
=\sum_{\alpha=1}^{k}g_{\alpha}^{2}(j)p_{+j}$

$0=\sum_{\alpha=1}^{k}f_{\alpha}(i)p_{i+}=\sum_{\alpha=1}^{k}g_{\alpha
}(j)p_{+j}$

$0=\sum_{\alpha=1}^{k}f_{\alpha}(i)f_{\beta}(i)p_{i+}=\sum_{\alpha=1}%
^{k}g_{\alpha}(j)g_{\beta}(j)p_{+j}$ \ \ for $\alpha\neq\beta.$\bigskip

In TCA, the parameters $(f_{\alpha}(i),g_{\alpha}(j),\delta_{\alpha})$ in (9)
satisfy: for $\alpha,\beta=1,...,k$

$\delta_{\alpha}=\sum_{\alpha=1}^{k}|f_{\alpha}^{\ }|(i)p_{i+}=\sum_{\alpha
=1}^{k}|g_{\alpha}(j)|p_{+j}$

$0=\sum_{\alpha=1}^{k}f_{\alpha}(i)p_{i+}=\sum_{\alpha=1}^{k}g_{\alpha
}(j)p_{+j}$

$0=\sum_{\alpha=1}^{k}f_{\alpha}(i)\ sign(f_{\beta}(i))p_{i+}=\sum_{\alpha
=1}^{k}g_{\alpha}(j)\ sign(g_{\beta}(j))p_{+j}$\ \ for $\alpha>\beta.\bigskip$

The following result is essential in CA of extremely sparse contingency
tables.\bigskip

\textbf{Theorem 1}: (Benz\'{e}cri 1973, pp188-190): If $\delta_{1}=1$ in CA,
then the contingency table has two blocks diagonal structure.\bigskip

\textbf{Remark 2}: Benz\'{e}cri (1973, p.189-190) observed that
\textquotedblright for sparse data\textquotedblright\ \ it is rare to have
$\delta_{1}=1;$ but \textquotedblright\textit{not uncommon}\textquotedblright%
\ to have $\delta_{1}\geq0.837,$ then the structure of the contingency table
could be \textit{quasi-2-blocks diagonal. }For an example see Choulakian
(2021)\textit{.}

In the next subsection, we explain in what sense TCA is ROBUST\ compared to CA.

\subsection{The robustness of TCA}

TCA is computed in 3 steps iteratively. We detail the 1st iteration $\alpha
=1$\bigskip

$\mathbf{P}^{(\alpha)}=\mathbf{P}^{(1)}=(p_{ij}^{(1)}=p_{ij}-p_{i+}p_{+j}%
$)\ \ is the cross-covariance matrix

Step 1: computation

$\ \ \ \ \ \ \ \ \ \ \ \ \ \mathbf{u}_{1}=(u_{1}(j))=\arg\max_{\mathbf{u\in
}\left\{  -1,1\right\}  }||\mathbf{P}^{(1)}\mathbf{u}||_{1}$

$\ \ \ \ \ \ \ \ \ \ \ \ \ \mathbf{a}_{1}=\mathbf{P}^{(1)}\mathbf{u}_{1}$

$\ \ \ \ \ \ \ \ \ \ \ \ \mathbf{v}_{1}=sign(\mathbf{a}_{1})$

$\ \ \ \ \ \ \ \ \ \ \ \ \mathbf{b}_{1}=(\mathbf{P}^{(1)})^{\prime}%
\mathbf{v}_{1}$

$\ \ \ \ \ \ \ \ \ \ \ \ \mathbf{u}_{1}=sign(\mathbf{b}_{1})$

$\ \ \ \ \ \ \ \ \ \ \ \ \delta_{1}=\mathbf{v}_{1}^{\prime}\mathbf{P}%
^{(1)}\mathbf{u}_{1}$

Step 2: scaling

$\ \ \ \ \ \ \ \ \ \ \ \ \mathbf{f}_{1}=\mathbf{a}_{1}/(p_{i+})$

$\ \ \ \ \ \ \ \ \ \ \ \ \mathbf{g}_{1}=\mathbf{b}_{1}/(p_{+j})$

Step 3: residual matrix...

$\ \ \ \ \ \ \ \ \ \ \mathbf{P}^{(2)}=\mathbf{P}^{(1)}-f_{1}(i)g_{1}%
(j)/\delta_{1}$

$\alpha=\alpha+1$ \ go to Step 1\bigskip

The robustness of TCA follows from the following two observations that we
describe for the 1st iteration $\alpha=1$:

Fact 1: $p_{ij}^{(1)}=p_{ij}-p_{i+}p_{+j}$ \ \ is double-centered: $\sum_{i}$
$p_{ij}^{(1)}=\sum_{j}$ $p_{ij}^{(1)}=0.$

Fact 2: The coordinates of the principal axes (weights) $\mathbf{u}_{1}%
=(u_{1}(j))$ and $\mathbf{v}_{1}=(v_{1}(j))$ have values $1$ or $-1$.

These two facts imply that at iteration $\alpha=1$, the residual
(cross-covariance) matrix is divided into 4 \textit{balanced} quadrants
equally dispersed (not true in CA):%
\begin{align*}
\delta_{1}/4  &  =\sum_{i,j}(p_{ij}^{(1)}:\ v_{1}(i)=u_{1}(j)=1)\\
&  =\sum_{i,j}(p_{ij}^{(1)}:\ v_{1}(i)=u_{1}(j)=-1)\\
&  =\sum_{i,j}(p_{ij}^{(1)}:\ v_{1}(i)=-u_{1}(j)=1)\\
&  =\sum_{i,j}(p_{ij}^{(1)}:\ v_{1}(i)=-u_{1}(j)=-1).
\end{align*}

From which we define 5 QSR (quality of signs of residuals) indices for each
principal dimension: 1 global, 2 attractive (positive) and 2 repulsive (negative).

For $\alpha=1...k$

$QSR_{\alpha}=\frac{\delta_{\alpha}}{\sum_{i,j}|p_{ij}^{(\alpha)}|}$

$QSR_{\alpha}(v_{\alpha}(i)=u_{\alpha}(j)=1)=\frac{\delta_{\alpha}/4}%
{\sum_{(i,j:v_{\alpha}(i)=u_{\alpha}(j)=1)}|p_{ij}^{(\alpha)}|}$

$QSR_{\alpha}(v_{\alpha}(i)=u_{\alpha}(j)=-1)=\frac{\delta_{\alpha}/4}%
{\sum_{(i,j:v_{\alpha}(i)=u_{\alpha}(j)=-1)}|p_{ij}^{(\alpha)}|}$

$QSR_{\alpha}(v_{\alpha}(i)=-u_{\alpha}(j)=1)=-\frac{\delta_{\alpha}/4}%
{\sum_{(i,j:v_{\alpha}(i)=-u_{\alpha}(j)=1)}|p_{ij}^{(\alpha)}|}$

$QSR_{\alpha}(v_{\alpha}(i)=-u_{\alpha}(j)=-1)=-\frac{\delta_{\alpha}/4}%
{\sum_{(i,j:v_{\alpha}(i)=-u_{\alpha}(j)=-1)}|p_{ij}^{(\alpha)}|}$\ 

\ \ \ \ \ \ \ \ \ \ 

The 5 QSR indices satisfy\bigskip

\textbf{Lemma 1}: For $\mathbf{P}^{(\alpha)}$ for $\alpha=1,...,k$ and
rank($\mathbf{P}^{(1)})=k$

a) $-1\leq QSR\leq1$

b) For $\alpha=1,...,k-1,$ $QSR_{\alpha}=1$ if and only if

$QSR_{\alpha}(v_{\alpha}(i)=u_{\alpha}(j)=1)=QSR_{\alpha}(v_{\alpha
}(i)=u_{\alpha}(j)=-1)=$

$-QSR_{\alpha}(v_{\alpha}(i)=-u_{\alpha}(j)=1)=-QSR_{\alpha}(v_{\alpha
}(i)=-u_{\alpha}(j)=-1)=1.$

c) For $\alpha=k,$ $QSR_{\alpha}=1.\bigskip$

In the next section, we analyze-visualize the religious data set by CA and TCA
using two R packages: \textit{ca} by Greenacre, Nenadic and Friendly (2020)
and \textit{TaxicabCA} by Allard and Choulakian (2019).

\section{CA and TCA maps}

The data set, constructed by (Sah and Fokou\'{e}, 2019), concerns $I=590$
fragments of texts extracted from English translations of 8 religious-sacred
scripts: the first 4 of them from the ancient Bible and the rest from
\ extreme Orient countries Tibet (B), China (T) and India (Y, U):\bigskip

\textit{Book of Proverb} (P) composed of 31 chapters

\textit{Book of Ecclesiastes} (E) composed of 12 chapters

\textit{Book of Ecclesiasticus} (e) composed of 50 chapters

\textit{Book of Wisdom} (W) composed of 19 chapters

\textit{Four Noble Truth of Buddhism} (B) composed of 45 chapters

\textit{Tao Te Ching} (T) composed of 81 chapters

\textit{Yogasutras} (Y) composed of 189 chapters

\textit{Upanishads} (U) composed of 162 chapters\bigskip

\textbf{Remark 3:}

a) A symbol such as (P) represents a chapter from the \textit{Book of
Proverb}. The 8 letters (P, E, e, W, B, T, Y, U) in the parentheses will be
used to represent the rows of the data set on the CA and TCA maps.

b) The total number of ORIENTAL\ chapters is 478 (number of words counted is
30252), which is much larger than (almost equal to) the total number of
Biblical chapters 112 (number of words counted is 30355). This means that on
the average the chosen fragments of ORIENTAL\ books are much sparser than the
chosen fragments from the BIBLICAL\ books.

c) The \textit{apparent \%} of zero cells is 99.14\% based on the initial data
set of size $590\times8265$. While by Benz\'{e}cri's principle of
distributional equivalence property (see Definitions 1 and 2), the
\textit{real \%} of zero cells is 98.65\% based on the merged data set of size
$589\times4864$. Note that the marginal count of the words in the chosen
fragment of ch 14 of the B book is null; so in CA and TCA this row is
eliminated. So we consider the data set as extremely sparse.

\subsection{CA map}

Figure 1 displays the CA map, where the \% of explained inertia is 1.1\%
$(06+0.5)$ considered too small, and the Buddhism (B) chapters dominate the
first and the second principal dimensions. However, examining the first four
dispersion values (0.80, 0.72, 0.71 and 0.70), we see that the first
dispersion value 0.80 is quite near to Benz\'{e}cri's advocated lower bound of
0.837 as stated in Remark 2. \ So we can guess that the structure of the
contingency table could be \textit{quasi-2-blocks diagonal.}%

\begin{figure}[h]
\label{fig:MagEngT}
\centering 
\includegraphics[scale=1.3]{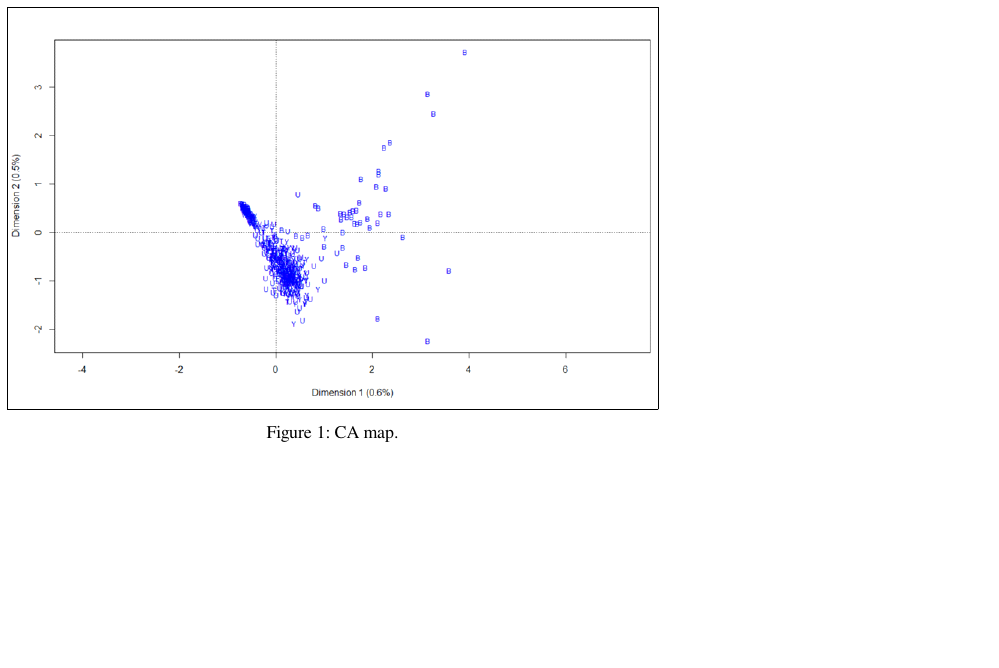}  
\end{figure}

\subsection{TCA maps}

Figure 2\ displays the TCA map of 589 chapters, where we notice that: a) The
first principal dimension impressively separates the EASTERN books and the
Bible books. This confirms Benz\'{e}cri's observation that the data set has
quasi-2-blocks structure. The Eastern books are much more dispersed than the
Bible books, because of Remark 3b. b) Concerning the interpretation of the
second principal dimension, we observe: on the right clear separation of the
books (B,T) and the books (Y,U); and on the left, the separation of the book
(W) from the rest.%

\begin{figure}[h]
\label{fig:MagEngT}
\centering 
\includegraphics[scale=1]{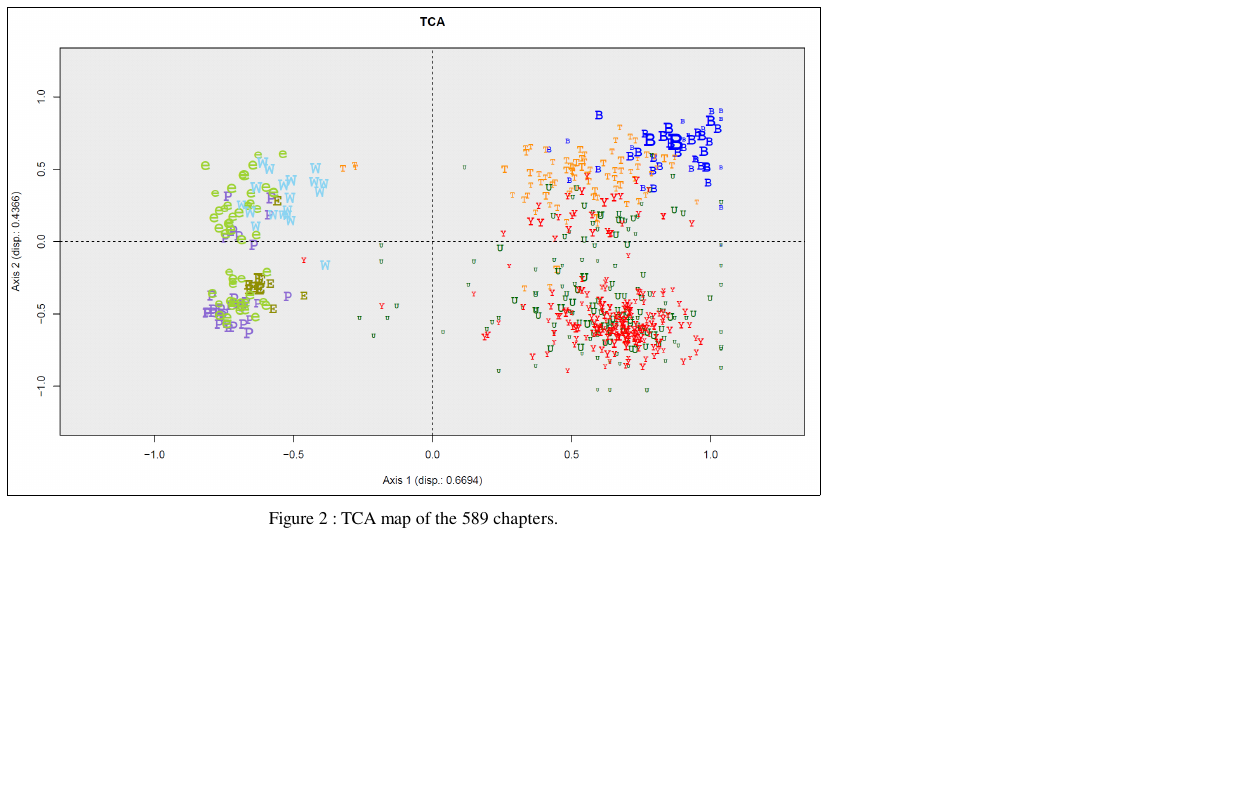}  
\end{figure}

Figure 3 displays the TCA map of the words, where we observe that: a) The
EASTERN books are positively-attractively associated with the words (pain,
view, mental, bodily, external, perception, effort, consciousness, eternal,
mystical, spiritual, psychical, powers) representing mainly \textit{intangible
concepts}; b) The BIBLE books are positively-attractively associated with the
words (enemies, hands, mercy, judgement, grace, justice, gold, lips, ways,
tongue mouth, wicked, house, poor) representing mainly \textit{tangible
objects}.%

\begin{figure}[h]
\label{fig:MagEngT}
\centering 
\includegraphics[scale=1.3]{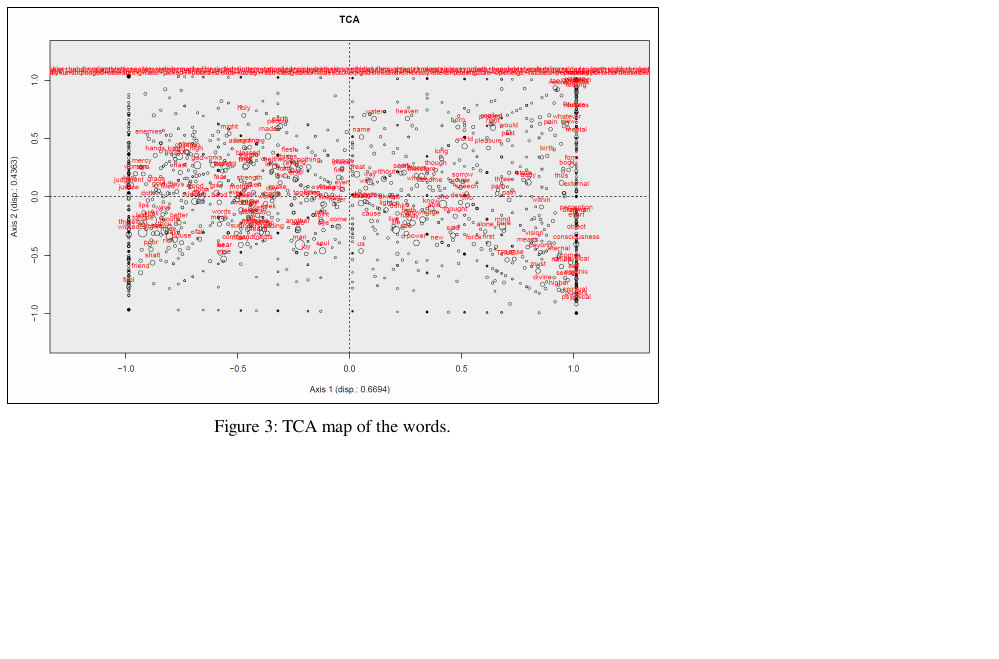}  
\end{figure}

Table 1 displays the five QSR indices for the first four principal dimensions.
QSR$_{1}=38.3\%$ is clearly different from the rest. The first two columns
show that the attractive associations (27.7\% and 31.1\%) are much smaller in
absolute values than the repulsive associations $(-59.1\%,$ $-51.9\%)$; which
implies the role played by the extreme sparsity of the contingency table thus
exhibiting the quasi-2-blocks diagonal structure.%

\begin{tabular}
[c]{lllll}%
\multicolumn{5}{l}{\textbf{Table 1: QSR (\%) for the first 4 principal
dimensions.}}\\\hline\hline
& TCA of $(n_{ij})$ &  &  & \\
Axis $\alpha$ & $QSR_{\alpha}(V_{+}U_{+},V_{-}U_{-})$ & $QSR_{\alpha}%
(V_{-}U_{+},V_{+}U_{-})$ & $QSR_{\alpha}$ & $\delta_{\alpha}$\\\hline
\textbf{1} & \textbf{(27.7,\ 31.1)} & \textbf{(-59.1,\ -51.9)} & \textbf{38.3}
& \textbf{0.669}\\
2 & (21.3, 21.1) & (-31.8, -33.7) & 25.7 & 0.437\\
3 & (23.3,\ 16.0) & (-30.2,\ -36.5) & 24.1 & 0.431\\
4 & (18.5, 19.4) & (-28.0, -27.8) & 22.6 & 0.391\\\hline
\end{tabular}

\section{Conclusion}

There are two main approaches for dimension reduction of high-dimensional data
sets SVD-like or MDS-like. MDS-like methods are distance based, often
Euclidean. Recently developed methods (t-SNE, UMAP, ...) in machine learning
are MDS-like, and they are characterized by the following two general points:
First, the pairwise distances computed in the high-dimensional spaces take
into consideration the nonlinear (local and global) structures, thus
emphasizing or balancing one aspect over the other; second, a loss-function is
minimized to compute the pairwise distances in the low-dimensional spaces; see
for example Wang et al. (2021). CA can be interpreted both as a generalized
SVD decomposition or a MDS-like method. While TCA is based on TSVD without a
loss function exactly imitating-generalizing Hotelling (1933)'s PCA approach.

The topic of interpretability being central in machine learning nowadays, see
for instance (Allen, Gan, Zheng 2023), this paper shows that our approach
based on CA framework is preferable to the approaches used in (Ma, Sun and Zou
2022): Figure 2 of this paper is very similar to Figure 3c (meta-spec) in (Ma,
Sun and Zou 2022); however the words embedded in Figur 3 of this paper brings
further interpretable details, which are missing in the methods discussed in
(Ma, Sun and Zou 2022) paper.

This paper showed on a case study that the joint use of CA and TCA are valid
visualization methods for sparse or extremely sparse contingency tables, and
they are worth trying.

\bigskip
\begin{verbatim}
\bigskip Acknowledgements
\end{verbatim}

Choulakian's research has been supported by NSERC of Canada.\bigskip

\textbf{References}

Allard J, Choulakian V (2019) \textit{Package TaxicabCA in R}

Allen GI, Gan L, Zheng L (2023) Interpretable machine learning for discovery:
Statistical challenges \& opportunities. https://arxiv.org/pdf/2308.01475.pdf

Beh E, Lombardo R (2014) \textit{Correspondence Analysis: Theory, Practice and
New Strategies}. N.Y: Wiley

Belkin M, Niyogi P (2003). Laplacian eigenmaps for dimensionality reduction
and data representation. \textit{Neural Computation,} 15 (6), 1373-1396

Benz\'{e}cri JP (1973)\ \textit{L'Analyse des Donn\'{e}es: Vol. 2: L'Analyse
des Correspondances}. Paris: Dunod

Choulakian V (2006) Taxicab correspondence analysis. \textit{Psychometrika,}
71, 333-345

Choulakian V (2016) Matrix factorizations based on induced norms.
\textit{Statistics, Optimization and Information Computing}, 4, 1-14

Choulakian V (2017) Taxicab correspondence analysis of sparse contingency
tables. \textit{Italian Journal of Applied Statistics,} 29 (2-3), 153-179

Choulakian V (2021) Quantification of intrinsic quality of a principal
dimension in correspondence analysis and taxicab correspondence analysis.
Available on \textit{arXiv:2108.10685}

Choulakian V, Allard J, Mahdi S (2023) \ Taxicab correspondence analysis and
Taxicab logratio analysis: A comparison on contingency tables and
compositional data. \textit{Austrian Journal of Statistics,} 52, 39 -- 70

Donoho DL, Grimes C (2003). Hessian eigenmaps: Locally linear embedding
techniques for high-dimensional data. \textit{Proceedings of the National
Academy of Sciences,} 100 (10), 5591-5596

Fran\c{c}ois D, Wertz V, Verleysen M (2007) The concentration of fractional
distances. \textit{IEEE Transactions on Knowledge and Data Engineering}, 19
(7), 873--886.

Gower JC (1966) Some distance properties of latent root and vector methods
used in multivariate analysis. \textit{Biometrika} 53 (3), 325--338

Greenacre MJ (1984) \textit{Theory and Applications of Correspondence
Analysis}. Academic Press, London

Kruskal JB (1964) Nonmetric multidimensional scaling: A numerical method.
\textit{Psychometrika,} 29 (2), 115-129

Hotelling H (1933) Analysis of a complex of statistical variables into
principal components. \textit{Journal of Educational Psychology}, 24(6), 417-441

Lee JA, Verleysen M (2011) Shift-invariant similarities circumvent distance
concentration in stochastic neighbor embedding and variants. International
Conference on Computational Science, \textit{Procedia Computer Science} 4, 538--547

Ma R, Sun E, Zou J (2023) A spectral method for assessing and combining
multiple data visualizations. \textit{Nature Communications, }14(1):780 doi: 10.1038/s41467-023-36492-2

Ma R, Sun E, Zou J (2023) A spectral method for assessing and combining
multiple data visualizations. https://arxiv.org/pdf/2210.13711.pdf

Mallet-Gauthier S, Choulakian V (2015) Taxicab correspondence analysis of
abundance data in archeology: three case studies revisited.
\textit{Archeologia e Calcolatori}, 26, 77-94

McInnes L, Healy J, Melville J (2018). Umap: Uniform manifold approximation
and projection for dimension reduction. arXiv:1802.03426

Moon KR, van Dijk D, Wang Z, Gigante S, Burkhardt DB, Chen WS, Yim K, van den
Elzen A, Hirn MJ, Coifman DR, Natalia B Ivanova NB, Wolf G, Krishnaswamy S
(2019). Visualizing structure and transitions in highdimensional biological
data. \textit{Nature Biotechnology} 37 (12), 1482-1492

Roweis ST, Saul LK (2000) Nonlinear dimensionality reduction by locally linear
embedding. \textit{Science,} 290 (5500), 2323-2326

Sah P, Fokou\'{e} E (2019) What do asian religions have in common? an
unsupervised text analytics exploration. arXiv:1912.10847

Sammon JW (1969) A nonlinear mapping for data structure analysis. \textit{IEEE
Transactions on Computers,} 100 (5), 401-409

Scholkopf B, Smola A, Muller KR (1997) Kernel principal component analysis. In
\textit{International Conference on Artificial Neural Networks}, 583-588, Springer

Tenenbaum JB, Silva V, Langford JC (2000) A global geometric framework for
nonlinear dimensionality reduction. \textit{Science,} 290 (5500), 2319-2323

Torgerson W (1952) Multidimensional scaling: I. theory and method.
\textit{Psychometrika}, 17(4), 401--419

van der Maaten L, Hinton G (2008) Visualizing data using t-SNE.
\textit{Journal of Machine Learning Research} 9(Nov), 2579-2605

Wang Y, Huang H, Rudin C, Shaposhnik Y (2021) Understanding how dimension
reduction tools work: An empirical approach to deciphering t-sne, umap,
trimap, and pacmap for data visualization. \textit{Journal of Machine Learning
Research,} 22, 1-73

\end{document}